\documentclass{bmvc2k_arxiv}

\usepackage{booktabs}
\usepackage{gensymb}

\title{End-to-End Multi-View Lipreading}

\addauthor{Stavros Petridis}{stavros.petridis04@imperial.ac.uk}{1}
\addauthor{Yujiang Wang}{              yujiang.wang14@imperial.ac.uk }{1}
\addauthor{Zuwei Li}{               zuwei.li15@imperial.ac.uk }{1}
\addauthor{Maja Pantic}{              m.pantic@imperial.ac.uk}{12}

\addinstitution{
iBUG Group \\
Dept. Computing \\
Imperial College London \\
London, UK \\
}

\addinstitution{
EEMCS \\
University Of Twente \\
 Enschede, The Netherlands
}

\runninghead{Accepted to BMVC}{2017}

\begin{document}

\maketitle

\begin{abstract}
Non-frontal lip views contain useful information which can be used to enhance the performance of frontal view lipreading. However, the vast majority of recent lipreading works, including the deep learning approaches which significantly outperform traditional approaches, have focused on frontal mouth images. As a consequence, research on joint learning of visual features and speech classification from multiple views is limited. In this work, we present an end-to-end multi-view lipreading system based on Bidirectional Long-Short Memory (BLSTM) networks. To the best of our knowledge, this is the first model which simultaneously learns to extract features directly from the pixels and performs visual speech classification from multiple views and also achieves state-of-the-art performance. The model consists of multiple identical streams, one for each view, which extract features directly from different poses of mouth images. The temporal dynamics in each stream/view are modelled by a BLSTM and the fusion of multiple streams/views takes place via another BLSTM. An absolute average improvement of 3\% and 3.8\% over the frontal view performance is reported on the OuluVS2 database when the best two  (frontal and profile) and three views (frontal, profile, $45^\circ$) are combined, respectively. The best three-view model results in a 10.5\% absolute improvement over the current multi-view state-of-the-art performance on OuluVS2, without using external databases for training, achieving a maximum classification accuracy of 96.9\%.
\end{abstract}

%-------------------------------------------------------------------------
\section{Introduction}

Lipreading, also known as visual speech recognition, is the process of recognising speech by observing only the lip movements without having access to the audio signal. Several approaches have been presented \cite{zhao2009lipreading,Potamianos2003,Dupont2000,matthews2002extraction} which extract features from a mouth region of interest (ROI) and attempt to model their dynamics in order to recognise speech. Such a system has the potential to enhance acoustic speech recognition in noisy environments since the visual signal is not corrupted by acoustic noise and can also enable the use of silent interfaces. 

Recently, several deep learning approaches \cite{ninomiya2015integration,ngiam2011multimodal,petridis2016deep,sui2014extracting,chung2016lip,chung2016out} have been presented which replace the traditional feature extraction process and automatically extract features from the pixels. There are also a few end-to-end approaches which attempt to jointly learn the extracted features and perform visual speech classification \cite{petridis2017deepVisualSpeech,chung2016lipSentences,wand2016lipreading,assael2016lipnet}. This has given rise to a second generation of lipreading systems based on deep learning which significantly outperform the traditional approaches.

The vast majority of previous works has focused on frontal view lipreading. This is in contrast to evidence in the literature that human lip-readers prefer non-frontal views \cite{bauman1995analysis,lam2012}, probably due to lip protrusion and lip rounding being more pronounced. It is therefore reasonable to assume that non-frontal lip views contain useful information which can be used to enhance the performance of frontal view lipreading or in cases where frontal view of the mouth ROI is not available. This is especially true in realistic in-the-wild scenarios where the face is rarely frontal. However, research on multi-view lipreading has been very limited and mainly restricted to evaluating non-frontal views independently with few works combining up to two views.

Such a system would be useful in meeting rooms where multiple cameras can record the participants simultaneously. A car environment is another potential application where multiple cameras can be easily installed and multiple views are easily available. Finally, some new smartphones have dual frontal cameras so that is another potential scenario although the views will not be too far apart.

In this work, we present an end-to-end model which jointly learns to extract features directly from the pixels and performs visual speech classification from multiple views. To the best of our knowledge, this is the first end-to-end model which performs multi-view lipreading and also achieves state-of-the-art performance. The proposed model consists of multiple identical streams, one per view,  which extract features directly from the raw images. Each stream consists of an encoder which compresses the high dimensional input image to a low dimensional representation. The encoding layers in each stream are followed by a BLSTM which models the temporal dynamics. Finally,  the information of the different streams/views is fused  via a BLSTM which also provides a label for each input frame.

The second contribution of this work is an extensive comparison of all possible combinations up to five views. We evaluate the proposed model on the OuluVS2 database \cite{Anina2015}, which to the best of our knowledge is the only publicly available database containing 5 different views between 0\degree and 90\degree. We first perform an evaluation of each view independently and we see that the frontal and profile views are the best single views. The combination of these two views results in the best 2-view model which leads to a 3\% absolute increase in classification accuracy over the frontal view. The addition of the 45\degree view results in a further small improvement of 0.8\% over the best 2-view model. This means that multi-view lipreading can indeed enhance the performance of frontal view lipreading. However, the addition of more views (4 or 5) does not lead to any further improvements. We also show that non-frontal combinations like 30\degree or 45\degree and 90\degree outperform the frontal view which means that such combinations can be successfully used when frontal lip views are not available.  Finally, in all single-view or multi-view scenarios the proposed model improves the state-of-the-art performance and achieves a maximum classification accuracy of 96.9\%  which is the highest performance reported on the OuluVS2 database.

\begin{figure}[t]
  \centering
\includegraphics[width=0.8\linewidth, trim = {9cm 1cm 2cm 9cm},clip]{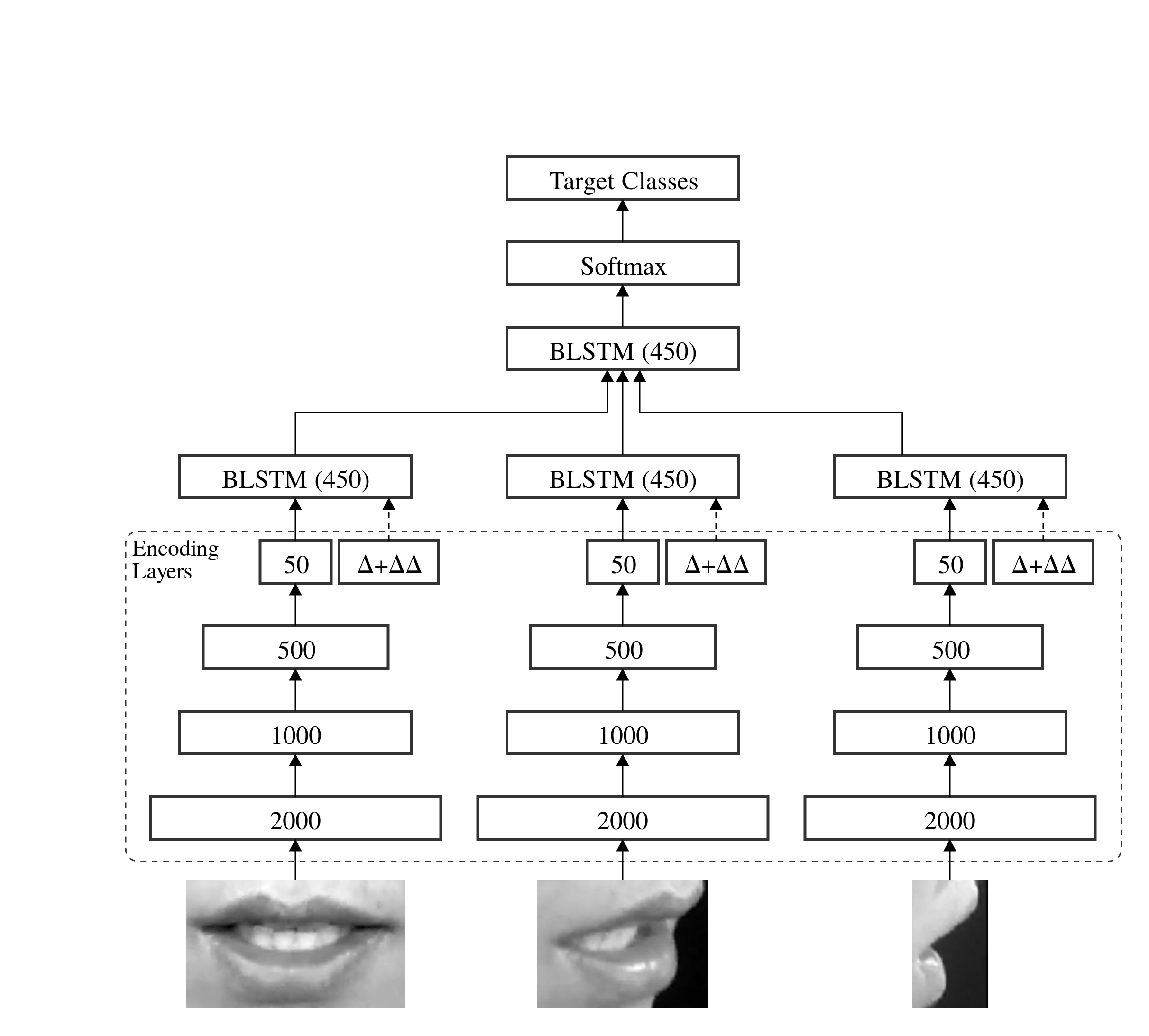}
  \caption{ Overview of the end-to-end visual speech recognition system. One stream per view is used for feature extraction directly from the raw images. Each stream consists of an encoder which compresses the high dimensional input image to a low dimensional representation. The $\Delta$ and $\Delta\Delta$ features are also computed  and appended to the bottleneck layer. The encoding layers in each stream are followed by a BLSTM which models the temporal dynamics. A BLSTM is used to fuse the information from all streams and provides a label for each input frame.}
\label{fig:3streamNetwork}
\end{figure}

\section{Related Work}

% One of the main reason why works on non-frontal and multi-view lipreading are limited is the lack of suitable databases. Only recently, few databases have been presented which also contain non-frontal poses but still most of them are restricted to two views and they are not publicly available.

\looseness - 1
Previous works can be grouped into three categories: 1) evaluation of different views, 2) pose-invariant lipreading and 3) multi-view lipreading. The first group contains works which evaluate non-frontal views individually and compare their performance with the frontal view.
It is not possible to draw a conclusion as to which is the most informative view since conflicting results are presented in the literature. Early works considered only frontal and profile views. Lucey and Potamianos \cite{lucey2006ProfVSfrontal} report a moderate performance degradation when profile view is used  compared to frontal view. On the other hand, Kumar et al. \cite{kumar2007Profile} show results where the profile view outperforms the frontal view. Recently, few studies considering more views have been published. Lan et al. \cite{Lan2012ViewInd} evaluated 5 different views, 0\degree, 30\degree, 45\degree, 60\degree and 90\degree, for lipreading and showed that their system performed best at 30\degree. The same 5 views have been evaluated on the OuluVS2 database and the results are still conflicting. The frontal view was found to be the best by Saitoh et al. \cite{saitoh2016concatenated}, the profile view by  Lee at al. \cite{lee2016multi}, the 30\degree view by Zimmermann et al. \cite{zimmermann2016visual} and the 60\degree view was found to be the best performing in \cite{ouluVS2Results}. The results presented by Saitoh et al. provide some insight as to why this may happen. Three different convolutional neural networks (CNNs), GoogLeNet, AlexNet and Network in Network, were trained on OuluVS2 using data augmentation. Each model led to different performance across the views. This is probably an indication that the best view depends on the model and maybe even the features used. This could explain the different conclusions reached by different studies.

Pose-invariant lipreading works belong to the second group. The main goal of such works is to reduce the impact of different poses as it is known that the performance decreases in mismatched train/test conditions, i.e., the classifier is trained and tested on different poses. There are two main approaches in the literature for pose-invariant lipreading. The first one trains classifiers using data from all available views in order to build a generic classifier \cite{lucey2008continuous}. The second approach applies a mapping to transform features from non-frontal views to the frontal view. Lucey et al. \cite{lucey2007Ext} apply a linear mapping to transform profile view features to frontal view features. This approach has been extended to mapping other views like 30\degree, 45\degree and 60\degree to the frontal view \cite{Estellers2011} or to the 30\degree view \cite{Lan2012ViewInd}. This approach makes it possible to collect a large amount of data on the optimal view and train a robust model. However, as the number of features to be generated by the linear mapping increases the performance is degraded \cite{lucey2007Ext}.

To the best of our knowledge, there are only three works which have attempted multi-view lipreading, i.e., using multiple views of the lips simultaneously. Lucey and Potamianos \cite{lucey2006ProfVSfrontal} concatenated discrete cosine transform features extracted from frontal and profile views and fed them to a Hidden Markov Model (HMM). The performance of the 2-view system outperformed the frontal view model. Lee et al. \cite{lee2016multi} experimented with different CNN architectures which takes as input multiple lip views. The CNNs were combined with LSTMs and trained end-to-end on the OuluVS2 database. They just tested the combination of all 5 views which turned out to be worse than the frontal and profile views, possibly due to lack of enough training data. Finally, Zimmermann et al. \cite{zimmermann2016visual} used principal componenent analysis (PCA) networks together with LSTMs and HMMs in order to combine multiple views from the OuluVS2 database. The only combination which outperformed the frontal view is frontal + 30\degree. Similarly to \cite{lee2016multi}, the combination of all views led to worse performance than most individual views.

\section{End-to-End Multi-view Lipreading}

The proposed deep learning system for multi-view lipreading is shown  in Fig. \ref{fig:3streamNetwork}. It consists of three identical streams which extract features directly from the raw input images. Each stream corresponds to one view and consists of two parts: an encoder and a BLSTM. The encoder follows a bottleneck architecture in order to compress the high dimensional input image to a low dimensional representation at the bottleneck layer.  The same architecture as in \cite{hinton2006reducing} is used, with 3 fully connected hidden layers of sizes 2000, 1000 and 500, respectively, followed by a linear bottleneck layer. The rectified linear unit is used as the activation function for the hidden layers. The $\Delta$ (first derivatives) and $\Delta\Delta$ (second derivatives) \cite{young2002htk} features are also computed, based on the bottleneck features, and they are appended to the bottleneck layer. In this way, during training we force the encoding layers to learn compact representations which are discriminative for the task at hand but also produce discriminative $\Delta$ and $\Delta\Delta$ features. This is in contrast to the traditional approaches which pre-compute the $\Delta$ and $\Delta\Delta$ features at the input level and as a consequence there is no control over their discriminative power. 

The second part is a BLSTM layer added on top of the encoding layers in order to model the temporal dynamics associated with each view. The BLSTM outputs of each stream are concatenated and fed to another BLSTM in order to fuse the information from all streams and model the temporal dynamics associated with all views. The output layer is a softmax layer which provides a label for each input frame. The entire system is trained end-to-end which enables the joint learning of features and classifier. In other words, the encoding layers learn to extract features from raw images which are useful for classification using BLSTMs. 

\subsection{Single Stream Training}

\noindent
\textbf{Initialisation:} First, each stream/view is trained independently. The encoding layers are pre-trained in a greedy layer-wise manner using Restricted Boltzmann Machines (RBMs) \cite{hinton2012practical}.  Since the input (pixels) is real-valued and the hidden layers are either rectified linear or linear (bottleneck layer) four Gaussian RBMs \cite{hinton2012practical}  are used. Each RBM is trained for 20 epochs with a mini-batch size of 100 and L2 regularisation coefficient of 0.0002 using contrastive divergence. The learning rate is fixed to 0.001  as suggested in \cite{hinton2012practical} when the visible/hidden units are linear.

\noindent \looseness - 1
\textbf{End-to-End Training:} Once the encoder has been pretrained then the BLSTM is added on top and its weights are initialised using glorot initialisation \cite{glorot2010understanding}.
The Adam training algorithm is used for end-to-end training with a mini-batch size of 10 utterances. The default learning rate of 0.001 led to unstable training so it was reduced to 0.0003. Early stopping with a delay of 5 epochs was also used in order to avoid overfitting and gradient clipping was applied to the LSTM layers. 

\subsection{Multi-Stream Training} 

\textbf{Initialisation:} Once the single streams have been trained then they are used for initialising the corresponding streams in the multi-stream architecture. Then another BLSTM is added on top of all streams in order to fuse the single stream outputs. Its weights are initialised using glorot initialisation.

\noindent
\textbf{End-to-End Training:} Finally, the entire network is trained jointly using Adam \cite{kingma2014adam}, 0.0001, to fine-tune the entire network. Early stopping and gradient clipping were also applied similarly to single stream training.

\section{Experiments}
\label{sec:exp}

\begin{table}[tb]
\renewcommand{\arraystretch}{1.1}
\caption{Size of mouth ROIs in pixels for each view.}
\label{tab:mouthROIsize}
\centering
\begin{tabular}{cccccc}
\toprule  Views & \multicolumn{1}{c}{0\degree}  & \multicolumn{1}{c}{30\degree} & \multicolumn{1}{c}{45\degree} & \multicolumn{1}{c}{60\degree} & \multicolumn{1}{c}{90\degree}  \\

\midrule Height/Width & 29/50 & 29/44 & 29/43 & 35/44 & 44/30 \\

\bottomrule

\end{tabular} 

\end{table}

The database used in this study is the OuluVS2 \cite{Anina2015} which to the best of our knowledge is the only publicly available database with 5 lip views between 0\degree and 90\degree. It contains 52 speakers saying 10 utterances, 3 times each, so in total there are 156 examples per utterance. The utterances are the following: ``Excuse me'', ``Goodbye'', ``Hello'', ``How are you'', ``Nice to meet you'', ``See you'', ``I am sorry'', ``Thank you'', ``Have a good time'', ``You are welcome''.  The mouth ROIs are provided and they are downscaled as shown in Table \ref{tab:mouthROIsize} in order to keep the aspect ratio of the original videos constant.

We first partition the data into training, validation and test sets. The protocol suggested by the creators of the OuluVS2 database is used \cite{saitoh2016concatenated} where 40 subjects are used for training and validation and 12 for testing. We randomly divided the 40 subjects into 35 and 5 subjects for training and validation purposes, respectively.  This means that there are 1050 training utterances, 150 validation utterances and 360 test utterances.

\begin{table}[tb]
\renewcommand{\arraystretch}{1.1}
\caption{Mean (standard deviation) single-view classification accuracy of 10 runs on the OuluVS2 database. The end-to-end model is trained using the established evaluation protocol where 40 subjects are used for training and validation and 12 for testing. $\dagger$ denotes that the difference with the frontal view mean classification accuracy is statistically significant.}
\label{tab:resultsSingleView}
\centering
\begin{tabular}{ccccccc}
\toprule  Views & \multicolumn{1}{c}{0\degree}  & \multicolumn{1}{c}{30\degree} & \multicolumn{1}{c}{45\degree} & \multicolumn{1}{c}{60\degree} & \multicolumn{1}{c}{90\degree}  \\

\midrule End-to-end Encoder + \\
BLSTM  & 91.8 (1.1) & 87.3$^\dagger$ (1.6) & 88.8$^\dagger$ (1.4) & 86.4$^\dagger$ (0.6) & 91.2 (1.3)  \\ 

\bottomrule

\end{tabular} 

\end{table}

\begin{table}[t]
\renewcommand{\arraystretch}{1.1}
\caption{Comparison of the maximum single-view classification accuracy over 10 runs with previous works on OuluVS2 database. All works follow the established evaluation protocol where 40 subjects are used for training and validation and 12 for testing. $^\ast$ In cross-view training, the model is first trained with data from all views and then fine-tuned with data from the corresponding view. $^{\ast\ast}$ These models are pretrained on the BBC dataset \cite{chung2016lip}, which is a large database, and then fine-tuned on OuluVS2.  DA: Data Augmentation, LVM: Latent Variable Models}
\label{tab:resultsSingleViewMax}
\centering
\begin{tabular}{ccccccc}
\toprule  Views & \multicolumn{1}{c}{0\degree}  & \multicolumn{1}{c}{30\degree} & \multicolumn{1}{c}{45\degree} & \multicolumn{1}{c}{60\degree} & \multicolumn{1}{c}{90\degree}  \\

\midrule CNN + DA \cite{saitoh2016concatenated}  & 85.6 & 82.5 & 82.5 & 83.3 & 80.3  \\ 
\midrule End-to-end CNN + LSTM \cite{lee2016multi} & 81.1 & 80.0 & 76.9 & 69.2 & 82.2 \\
\midrule CNN + LSTM, Cross-view Training $^\ast$ \cite{lee2016multi} & 82.8 & 81.1 & 85.0 & 83.6 & 86.4 \\
\midrule PCA Network + LSTM + GMM-HMM \cite{zimmermann2016visual} & 74.1 & 76.8 & 68.7 & 63.7 & 63.1 \\
\midrule Baseline: Raw Pixels + LVM  \cite{ouluVS2Results} & 73.0 & 75.0 & 76.0 & 75.0 & 70.0 \\
\midrule End-to-end Encoder + BLSTM (Ours) & 94.7 & 89.7 & 90.6 & 87.5 & 93.1  \\
\midrule CNN pretrained on BBC dataset + DA$^{\ast\ast}$ \cite{chung2016lip} & 93.2 & - & - & - & - \\
\midrule CNN pretrained on BBC dataset + DA + LSTM$^{\ast\ast}$ \cite{chung2016out} & 94.1 & - & - & - & - \\

\bottomrule

\end{tabular} 

\end{table}

The target classes are a one-hot encoding for the 10 utterances. Each frame is labelled based on the label of the utterance and the end-to-end model is trained with these labels. The model provides a label for each frame and the majority label over each utterance is used in order to label the entire sequence.

Since all experiments are subject independent we first need to reduce the impact of subject dependent characteristics. This is done by subtracting the mean image, computed over the entire utterance, from each frame. The next step is the normalisation of data. As recommended in \cite{hinton2012practical} the data should be z-normalised, i.e. the mean and standard deviation should be equal to 0 and 1 respectively, before training an RBM with linear input units. Hence, each image is z-normalised before pre-training the encoding layers. 

\looseness - 1
Due to randomness in initialisation, every time a deep network is trained the results vary. In order to present a more objective evaluation we run each experiment 10 times and we report the mean and standard deviation of classification accuracy on the utterance level.

The proposed model was developed in Theano \cite{Theano} using the Lasagne \cite{lasagne} library. The code and best models for the best view combinations are publicly available \footnote{https://ibug.doc.ic.ac.uk/resources/EndToEndLipreading/}.

\subsection{Results on OuluVS2}

\begin{table}[t]
\renewcommand{\arraystretch}{1.1}
\renewcommand{\tabcolsep}{3pt}
\caption{Mean (standard deviation) 2-view classification accuracy over 10 runs of the proposed end-to end model on the OuluVS2 database. Due to lack of space we present results for the 6 best 2-view combinations. $\dagger$ denotes that the difference with the frontal view mean classification accuracy is statistically significant.}
\label{tab:results2View}
\centering
\begin{tabular}{cccccccc}
\toprule  Views & \multicolumn{1}{c}{0\degree + 30\degree}  & \multicolumn{1}{c}{0\degree + 45\degree} & \multicolumn{1}{c}{0\degree + 60\degree} & \multicolumn{1}{c}{0\degree + 90\degree} & \multicolumn{1}{c}{30\degree + 90\degree} & 45\degree + 90\degree \\

\midrule End-to-end \\
Encoder + BLSTM  & 91.7 (0.8) & 93.6$^\dagger$ (0.7) & 92.0 (1.0) & 94.8$^\dagger$ (0.7) & 93.8$^\dagger$ (0.8) & 93.6$^\dagger$ (0.7)  \\

\bottomrule

\end{tabular} 

\end{table}

\begin{table}[t]
\renewcommand{\arraystretch}{1.1}
\caption{Comparison of the maximum 2-view classification accuracy over 10 runs with previous works on the OuluVS2 database. All approaches are trained using the established evaluation protocol where 40 subjects are used for training and validation and 12 for testing. Due to lack of space we present results for the 6 best 2-view combinations. }
\label{tab:results2ViewMax}
\centering
\begin{tabular}{cccccccc}
\toprule  Views & \multicolumn{1}{c}{0\degree + 30\degree}  & \multicolumn{1}{c}{0\degree + 45\degree} & \multicolumn{1}{c}{0\degree + 60\degree} & \multicolumn{1}{c}{0\degree + 90\degree} & \multicolumn{1}{c}{30\degree + 90\degree} & 45\degree + 90\degree \\

\midrule PCA Network  \\ 
+ LSTM + HMM \cite{zimmermann2016visual} & 82.9 & 73.9 & 73.1 & 72.7 & - & -  \\
\midrule End-to-end \\
Encoder + BLSTM & 93.3 & 95.0 & 93.1 & 96.7 & 95 & 94.4  \\

\bottomrule

\end{tabular} 

\end{table}

\subsubsection{Single-View Results}

In the single view scenario, we train and test models on data recorded from a single view. This means that only a single stream is used from the model shown in Fig. \ref{fig:3streamNetwork}. Table \ref{tab:resultsSingleView} shows the mean classification accuracy and standard deviation of the 10 models trained for each view. The best performance is achieved by the frontal and profile views followed by the 45\degree, 30\degree and 60\degree views.

In almost all previous works just a single accuracy value is provided, which is most likely the maximum performance achieved, with no standard deviation. In order to facilitate a fair comparison, we also provide the maximum performance achieved over the 10 runs in Table \ref{tab:resultsSingleViewMax} together with previous results. It is obvious that even our mean performance is consistently higher for all views than previous works, which do not use external data for training. When it comes to maximum performance the proposed end-to-end architecture sets the new state-of-the-art for all views. It is worth pointing out, that the proposed system outperforms even the CNN models \cite{chung2016lip}, \cite{chung2016out} trained using external data. Both models are pre-trained on the BBC dataset \cite{chung2016lip}, which is a large dataset, and fine-tuned on OuluVS2.

\subsubsection{Multi-view Results}

\begin{table}[t]
\renewcommand{\arraystretch}{1.1}
\caption{3-view classification accuracy over 10 runs on the OuluVS2 database of the proposed end-to-end model. The mean (standard deviation) and the maximum accuracy over 10 runs are presented. Due to lack of space we present results for the 4 best 3-view combinations. $\dagger$ denotes that the difference with the frontal view mean classification accuracy is statistically significant.}
\label{tab:results3View}
\centering
\begin{tabular}{cccccc}
\toprule  Views & \multicolumn{1}{c}{0\degree + 45\degree + 90\degree}  & \multicolumn{1}{c}{0\degree + 30\degree + 90\degree} & \multicolumn{1}{c}{30\degree + 45\degree + 90\degree} & \multicolumn{1}{c}{0\degree + 60\degree + 90\degree}  \\
\midrule Mean (st. dev.) & 95.6$^\dagger$ (0.5) & 95.2$^\dagger$ (0.5) & 94.8$^\dagger$ (0.8) & 94.8$^\dagger$ (1.0)   \\
\midrule Max & 96.9 & 96.1 & 95.8 & 96.4 \\
\bottomrule

\end{tabular} 

\end{table}

In the multi-view scenario, we train and test models on data recorded from multiple views. Table \ref{tab:results2View} shows the mean classification accuracy and standard deviation over 10 runs for combinations of two views. Due to lack of space we present only the six best combinations and the other four are shown in the supplementary material. The combination of frontal and profile views is the best 2-view combination resulting in a 3\% absolute increase in classification accuracy over the frontal view. The only other combinations that outperform the frontal one are the following: 0\degree + 45\degree, 45\degree + 90\degree, 30\degree + 90\degree. It is beneficial to combine a frontal (0\degree) or partially frontal (30\degree or even 45\degree) view,
which contains information mostly about the lips appearance, with a side (90\degree) or partially side (45\degree) which contains information mostly about lip protrusion. It is also evident that combining neighbouring views is not useful since there is not enough complementary information. Finally, it is worth pointing out that although a 60\degree view is a partially side view its combination with any other view is not beneficial. This is because it is the worst performing view as can be seen in Table \ref{tab:resultsSingleView}. 

Table \ref{tab:results2ViewMax} shows the maximum performance over the 10 runs for the 2-view combinations. To the best of our knowledge only Zimmerman et al. \cite{zimmermann2016visual} present results on 2-view lipreading on OuluVS2. It is clear that the proposed end-to-end model significantly outperforms \cite{zimmermann2016visual} and sets the new state-of-the-art maximum performance for all combinations of 2 views.

\looseness - 1
Table \ref{tab:results3View} shows the results of the four best 3-view combinations, the rest can be found in the supplementary material.  All four combinations outperform the frontal view performance up to 3.8\%, however only 0\degree + 45\degree + 90\degree results in a statistically significant improvement of 0.8\% over the best 2-view combination (frontal + profile). Results for the combinations of 4 views and all views can be found in the supplementary material. All combinations outperform the frontal view up to 3.8\%, however none of them results in a statistically significant improvement over the best 3-view combination. This reveals that 2-view lipreading leads to the biggest improvement over frontal view lipreading, and the addition of a third view offers a further small improvement but addition of more views does not really offer any performance benefit. 

We should also point out that the maximum performance achieved of 96.9\% is the highest classification accuracy reported on the short phrases part of OuluVS2, outperforming the previous state-of-the-art accuracy of 86.4\% achieved by \cite{lee2016multi} (see Table \ref{tab:resultsSingleViewMax}) when no external data are used. It also outperforms the state-of-the-art performance of 94.1\%, when external data are used for training,  reported in \cite{chung2016out} (see Table \ref{tab:results2ViewMax}).

\begin{figure}[t]
  \centering
\includegraphics[width=\linewidth]{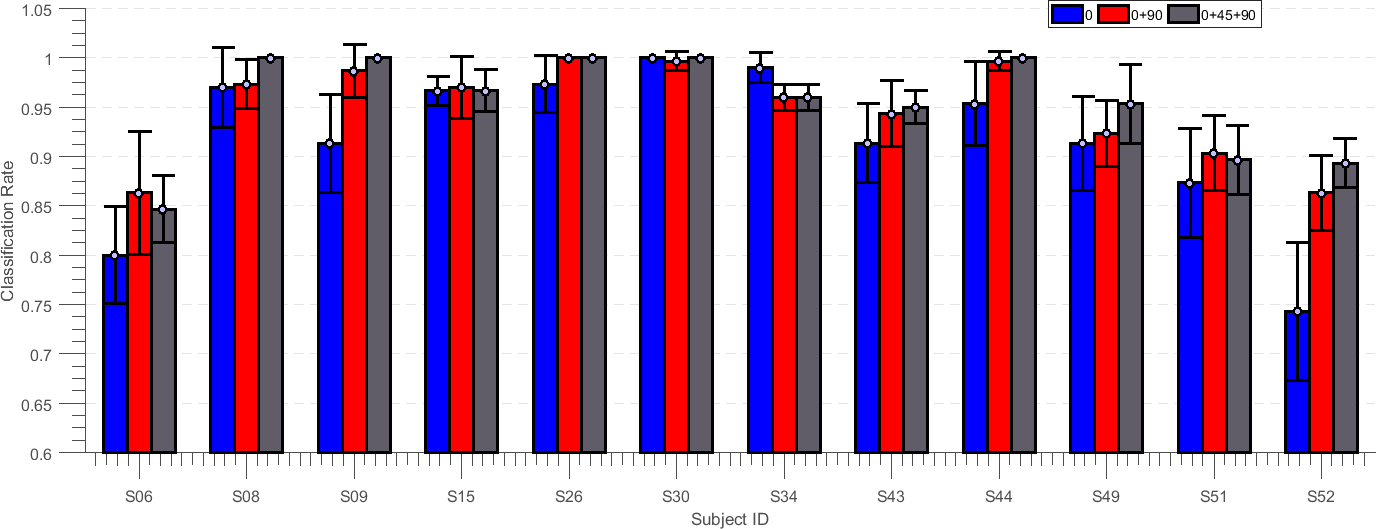}
  \caption{Mean classification accuracy and standard deviation per subject over 10 runs for the best performing views, frontal (blue), frontal + profile (red), frontal + 45\degree + profile (gray). }
\label{fig:clsfPerSubject}
\end{figure}

Fig. \ref{fig:clsfPerSubject} shows the classification accuracy per subject for the best single-view, 2-view and 3-view combinations, respectively. We observe that for all subjects except 34 the combinations of frontal and profile views and frontal + 45\degree + profile results in similar or better performance than the frontal view. It is clear that there is not a very large performance deviation across the different test subjects. All of them except 6 achieve a classification accuracy over 90\% with 5 of them (8, 9, 26, 30, 44) achieving 100\%.

\begin{figure}[t] 
  \subfigure[Frontal View]{% 
    \includegraphics[width=0.3\linewidth]{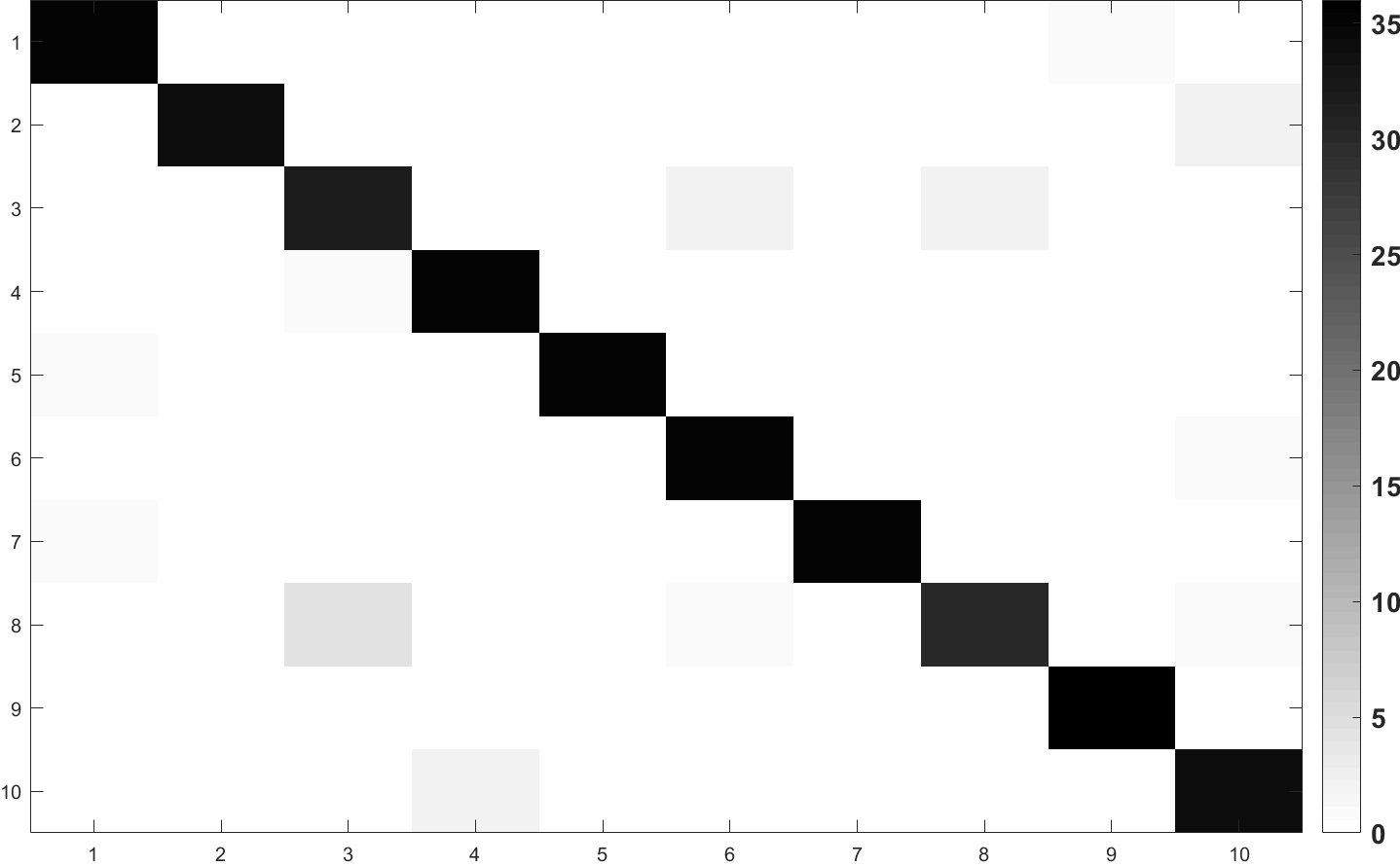} \label{fig:confMatFrontal} 
  } 
  \quad 
  \subfigure[Frontal + Profile Views]{% 
    \includegraphics[width=0.3\linewidth]{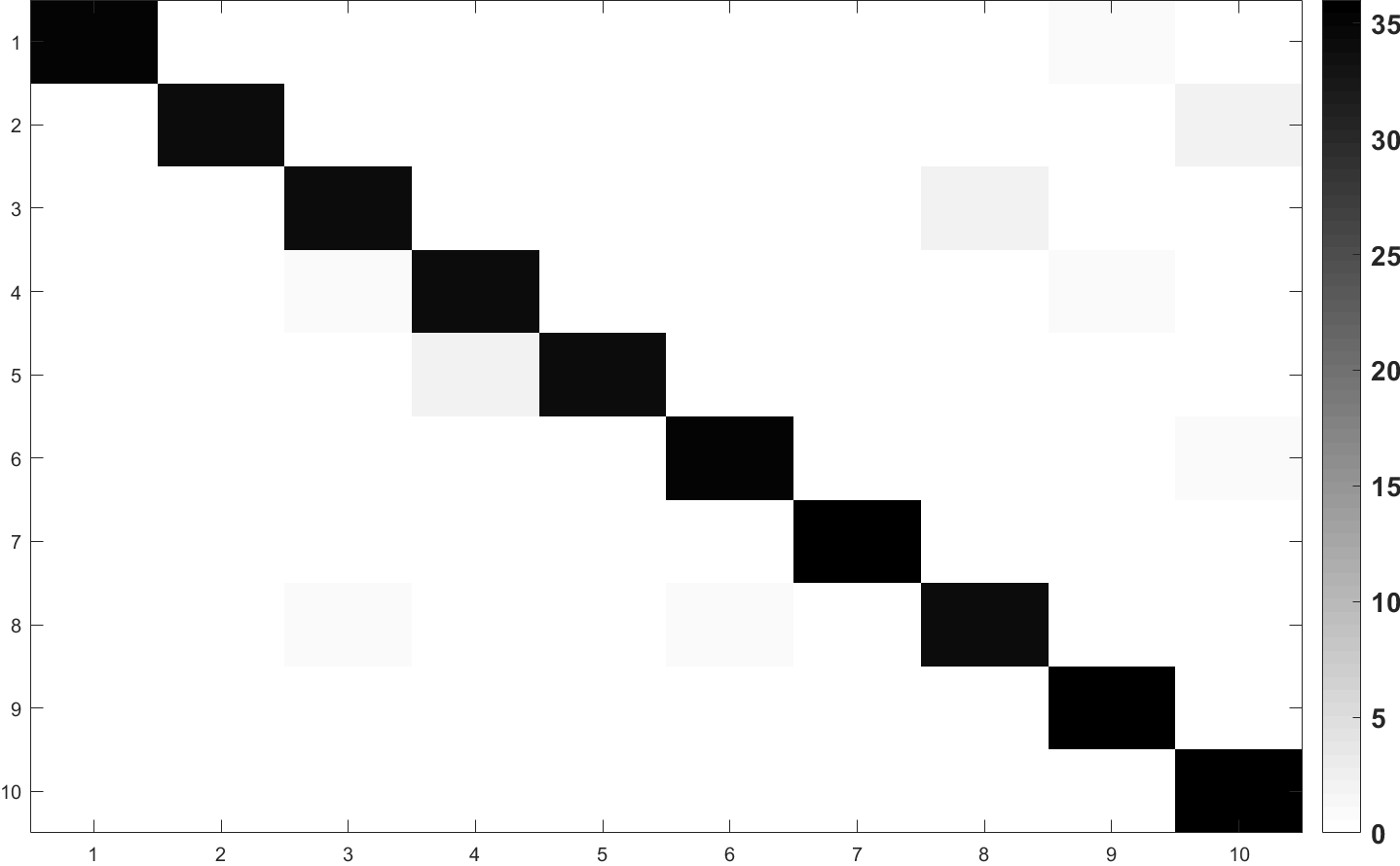} \label{fig:confMatFrontalProfile}
  } 
 \subfigure[Frontal + 45\degree + Profile Views]{% 
    \includegraphics[width=0.3\linewidth]{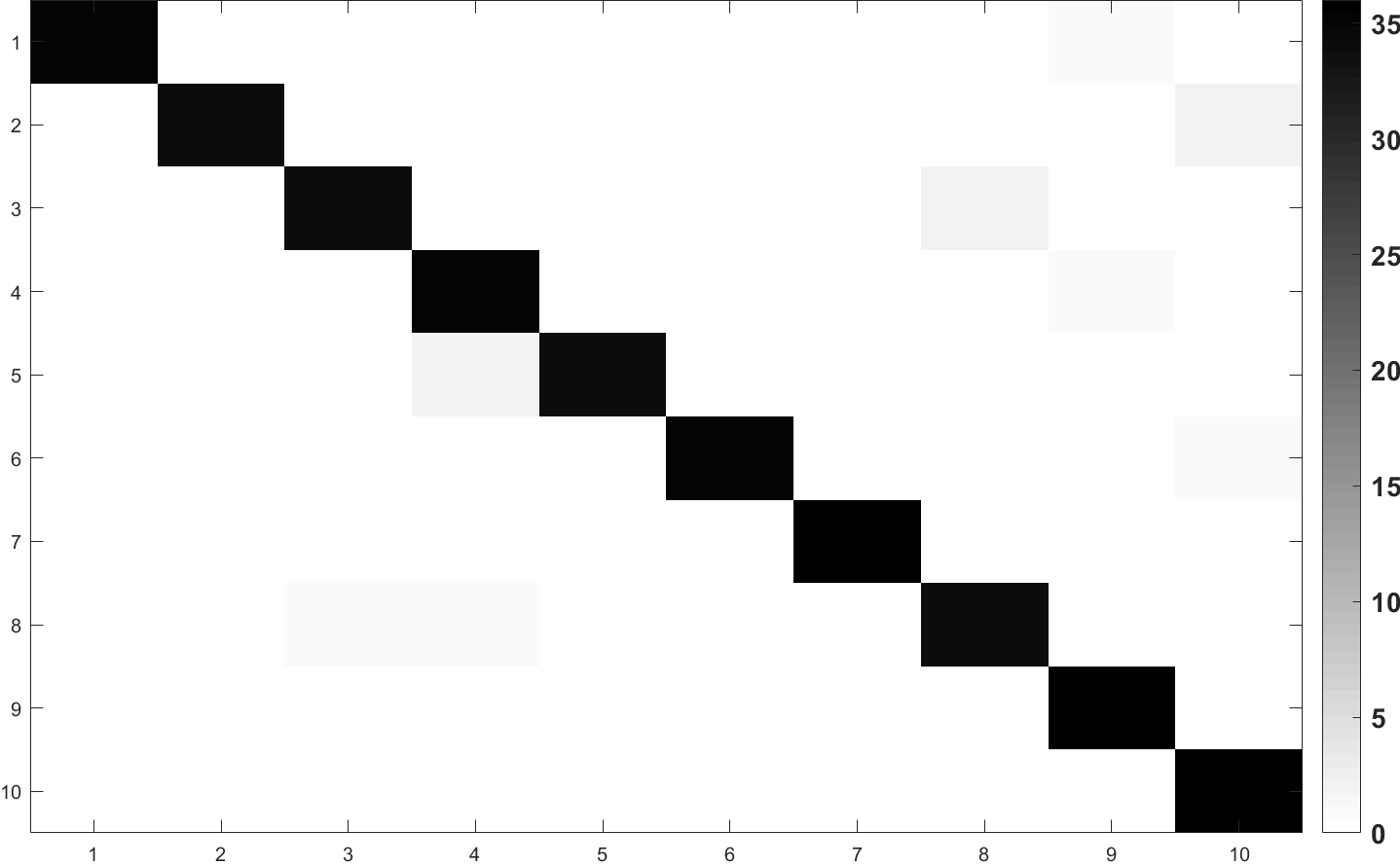} \label{fig:confMatFrontal45Profile}
  } 
  \caption{Confusion matrices for the best performing single-view, 2-view and 3-view models. The numbers correspond to the phrases in section \ref{sec:exp}. There are 12 subjects on the test set so there are 36 utterances per phrase. }\label{fig:confMat}
\end{figure}

Fig. \ref{fig:confMat} shows the confusion matrices for the best single-view, 2-view and 3-view combinations, respectively. It is clear that the number of confusions is reduced when multiple views are considered. The most common confusion pair in the single view is between ``Hello'' (3rd phrase) and ``Thank you'' (8th phrase) which is consistent with confusions presented in \cite{petridis2017deepVisualSpeech,lee2016multi}. This is followed by confusions between ``See you'' (6th phrase) and ``Hello'', ``Goodbye'' (2nd phrase) and ``You are welcome'' (10th phrase), ``You are welcome'' and ``How are you'' (4th phrase). 

Finally, we should also mention that we experimented with CNNs for the encoding layers but this led to worse performance than the proposed system. Chung and Zisserman  \cite{chung2016lip} report that it was not possible to train a CNN on OuluVS2 without the use of external data. Similarly, Saitoh et al. \cite{saitoh2016concatenated} report that they were able to train CNNs on OuluVs2 only after data augmentation was used. This is likely due to the rather small training set. We also experimented with data augmentation which improved the performance but did not exceed the performance of the proposed system.

\section{Conclusion}
In this work, we present an end-to-end multiview lipreading system which 
jointly learns to extract features directly from the pixels and performs classification using BLSTM networks. Results on the OuluVS2 demonstrate that the combination of multiple views is indeed beneficial for lipreading. The combination of the frontal and profile views leads to significant improvement over the frontal view. Further addition of the 45\degree view leads to a further small increase in performance. However, addition of more views does not lead to any further improvement. We have also demonstrated that combinations of non-frontal views, like 30\degree or 45\degree and 90\degree, can outperform the frontal view which is useful in cases where frontal lip views are not available.
The proposed model achieves state-of-the-art performance on the OuluVS2 without using external data for training or even data augmentation. However, we should stress that the provided mouth ROIs are well cropped and this might not be the case when automatic tools for mouth ROI detection are used. It is well known that the accuracy of automatic detectors might degrade with non-frontal views so it would be interesting to investigate the impact of automatic mouth ROI cropping on multi-view lipreading. Finally, the model can be easily extended to multiple streams so we are planning to add an audio stream in order to evaluate its performance on audiovisual multi-view speech recognition.

\section{Acknowledgements}
This work has been funded by the European Community Horizon
2020 under grant agreement no. 645094 (SEWA).

\appendix

\setcounter{table}{0}
\renewcommand{\thetable}{A\arabic{table}}

\section{Supplementary Material}

Results that were not included in the paper due to lack of space are included here.

\subsection{2-view results }

The 4 worst 2-view combinations are shown in Table \ref{tab:results2View}. As can be seen all combinations result in the same performance as the frontal view, none of the differences is statistically significant.

\begin{table}[tbh]
\renewcommand{\arraystretch}{1.1}
\renewcommand{\tabcolsep}{3pt}
\caption{Mean (standard deviation) and maximum classification accuracy of the 2-view combinations over 10 runs of the proposed end-to end model on the OuluVS2 database. This Table presents results for the 4 2-view combinations not included in the paper. $\dagger$ denotes that the difference with the frontal view mean classification accuracy is statistically significant.}
\label{tab:results2View}
\centering
\begin{tabular}{cccccc}
\toprule  Views & \multicolumn{1}{c}{45\degree + 60\degree}  & \multicolumn{1}{c}{30\degree + 45\degree} & \multicolumn{1}{c}{30\degree + 60\degree} & \multicolumn{1}{c}{60\degree + 90\degree} \\

\midrule Mean (st. dev.)  & 91.5 (0.7) & 91.4 (0.9) & 91.4 (0.8) & 91.1 (0.9)  \\
\midrule Max & 92.8 & 92.8 & 93.1 & 92.5 \\

\bottomrule

\end{tabular} 

\end{table}

\begin{table}[tbh]
\renewcommand{\arraystretch}{1.1}
\caption{3-view classification accuracy on the OuluVS2 database of the proposed end-to-end model. The mean (standard deviation) and the maximum accuracy over 10 runs are presented. $\dagger$ denotes that the difference with the frontal view mean classification accuracy is statistically significant.}
\label{tab:results3View1}
\centering
\begin{tabular}{ccccc}
\toprule  Views & \multicolumn{1}{c}{45\degree + 60\degree + 90\degree}  & \multicolumn{1}{c}{30\degree + 60\degree + 90\degree} & \multicolumn{1}{c}{0\degree + 45\degree + 60\degree}  \\
\midrule Mean (st. dev.) & 93.8$^\dagger$ (1.1) & 93.8$^\dagger$ (1.0) & 93.8$^\dagger$ (0.6)   \\
\midrule Max & 95.3 & 95.0 & 95.0 \\
\bottomrule

\end{tabular} 

\end{table}

\begin{table}[t]
\renewcommand{\arraystretch}{1.1}
\caption{3-view classification accuracy on the OuluVS2 database of the proposed end-to-end model. The mean (standard deviation) and the maximum accuracy over 10 runs are presented. $\dagger$ denotes that the difference with the frontal view mean classification accuracy is statistically significant.}
\label{tab:results3View2}
\centering
\begin{tabular}{ccccc}
\toprule  Views & \multicolumn{1}{c}{0\degree + 30\degree + 45\degree}  & \multicolumn{1}{c}{0\degree + 30\degree + 60\degree} & \multicolumn{1}{c}{30\degree + 45\degree + 60\degree}  \\
\midrule Mean (st. dev.) & 93.3$^\dagger$ (0.8) & 92.8 (0.9) & 92.3 (0.6)   \\
\midrule Max & 94.7 & 94.7 & 93.1 \\
\bottomrule

\end{tabular} 

\end{table}

\begin{table}[h!]
\renewcommand{\arraystretch}{1.1}
\caption{4-view classification accuracy on the OuluVS2 database of the proposed end-to-end model. The mean (standard deviation) and the maximum accuracy over 10 runs are presented. $\dagger$ denotes that the difference with the frontal view mean classification accuracy is statistically significant.}
\label{tab:results4View1}
\centering
\begin{tabular}{cccc}
\toprule  Views & \multicolumn{1}{c}{0\degree + 30\degree + 45\degree + 90\degree}  & \multicolumn{1}{c}{0\degree + 30\degree + 45\degree + 60\degree} & \multicolumn{1}{c}{0\degree + 30\degree + 60\degree + 90\degree}  \\
\midrule Mean (st. dev.) & 95.6$^\dagger$ (0.9) & 93.3$^\dagger$ (0.9) & 95.1$^\dagger$ (0.8)   \\
\midrule Max & 97.2 & 94.4 & 96.9 \\
\bottomrule

\end{tabular} 

\end{table}

\begin{table}[h!]
\renewcommand{\arraystretch}{1.1}
\caption{4-view classification accuracy on the OuluVS2 database of the proposed end-to-end model. The mean (standard deviation) and the maximum accuracy over 10 runs are presented.  $\dagger$ denotes that the difference with the frontal view mean classification accuracy is statistically significant.}
\label{tab:results4View2}
\centering
\begin{tabular}{ccc}
\toprule  Views &  \multicolumn{1}{c}{0\degree + 45\degree + 60\degree + 90\degree} & \multicolumn{1}{c}{30\degree + 45\degree + 60\degree + 90\degree}  \\
\midrule Mean (st. dev.) & 95.1$^\dagger$ (0.6) & 94.4$^\dagger$ (0.9)   \\
\midrule Max & 95.8 & 95.8 \\
\bottomrule

\end{tabular} 

\end{table}

\begin{table}[t]
\renewcommand{\arraystretch}{1.1}
\caption{5-view classification accuracyon the OuluVS2 database of the proposed end-to-end model. The mean (standard deviation) and the maximum accuracy over 10 runs are presented. $\dagger$ denotes that the difference with the frontal view mean classification accuracy is statistically significant.}
\label{tab:results5View}
\centering
\begin{tabular}{cc}
\toprule  Views &  All views \\
\midrule Mean (st. dev.) & 95.1$^\dagger$ (0.5)   \\
\midrule Max & 95.8  \\
\bottomrule

\end{tabular} 

\end{table}

\subsection{3-view results}

The 6 3-view combinations not included in the paper are shown in Tables \ref{tab:results3View1} and \ref{tab:results3View2}. All combinations except 0\degree + 30\degree + 60\degree and 30\degree + 45\degree + 60\degree outperform the frontal view performance but not the best 2-view combination 0\degree + 90\degree.

\subsection{4-view results}

The 4-view combinations are shown in Tables \ref{tab:results4View1} and \ref{tab:results4View2}. All of them result in a statistically significant improvement over the frontal view but offer no improvement over the best 3-view model 0\degree + 45\degree + 90\degree.

\subsection{5-view results }

The combination of all views is shown in Table \ref{tab:results5View}. It outperforms the frontal view performance but not the best 3-view model 0\degree + 45\degree + 90\degree.

\bibliography{egbib}

\begin{thebibliography}{34}
\providecommand{\natexlab}[1]{#1}
\providecommand{\url}[1]{\texttt{#1}}
\expandafter\ifx\csname urlstyle\endcsname\relax
  \providecommand{\doi}[1]{doi: #1}\else
  \providecommand{\doi}{doi: \begingroup \urlstyle{rm}\Url}\fi

\bibitem[The()]{Theano}
\url{http://deeplearning.net/software/theano/index.html/}.

\bibitem[las()]{lasagne}
\url{https://lasagne.readthedocs.io/en/latest/}.

\bibitem[oul()]{ouluVS2Results}
\url{http://www.ee.oulu.fi/research/imag/OuluVS2/preliminary.html}.

\bibitem[Anina et~al.(2015)Anina, Zhou, Zhao, and Pietik{\"a}inen]{Anina2015}
I.~Anina, Z.~Zhou, G.~Zhao, and M.~Pietik{\"a}inen.
\newblock Ouluvs2: A multi-view audiovisual database for non-rigid mouth motion
  analysis.
\newblock In \emph{IEEE FG}, pages 1--5, 2015.

\bibitem[Assael et~al.(2016)Assael, Shillingford, Whiteson, and
  de~Freitas]{assael2016lipnet}
Y.~M. Assael, B.~Shillingford, S.~Whiteson, and N.~de~Freitas.
\newblock Lipnet: Sentence-level lipreading.
\newblock \emph{arXiv preprint arXiv:1611.01599}, 2016.

\bibitem[Bauman and Hambrecht(1995)]{bauman1995analysis}
S.~L. Bauman and G.~Hambrecht.
\newblock Analysis of view angle used in speechreading training of sentences.
\newblock \emph{American journal of audiology}, 4\penalty0 (3):\penalty0
  67--70, 1995.

\bibitem[Chung and Zisserman(2016{\natexlab{a}})]{chung2016lip}
J.~S. Chung and A.~Zisserman.
\newblock Lip reading in the wild.
\newblock In \emph{Asian Conference on Computer Vision}, pages 87--103.
  Springer, 2016{\natexlab{a}}.

\bibitem[Chung and Zisserman(2016{\natexlab{b}})]{chung2016out}
J.~S. Chung and A.~Zisserman.
\newblock Out of time: automated lip sync in the wild.
\newblock In \emph{Workshop on Multiview Lipreading, Asian Conference on
  Computer Vision}, pages 251--263. Springer, 2016{\natexlab{b}}.

\bibitem[Chung et~al.(2017)Chung, Senior, Vinyals, and
  Zisserman]{chung2016lipSentences}
J.~S. Chung, A.~Senior, O.~Vinyals, and A.~Zisserman.
\newblock Lip reading sentences in the wild.
\newblock In \emph{The IEEE Conference on Computer Vision and Pattern
  Recognition (CVPR)}, July 2017.

\bibitem[Dupont and Luettin(2000)]{Dupont2000}
S.~Dupont and J.~Luettin.
\newblock Audio-visual speech modeling for continuous speech recognition.
\newblock \emph{IEEE Trans. on Multimedia}, 2\penalty0 (3):\penalty0 141--151,
  2000.

\bibitem[Estellers and Thiran(2011)]{Estellers2011}
V.~Estellers and J.~P. Thiran.
\newblock Multipose audio-visual speech recognition.
\newblock In \emph{European Signal Processing Conference}, pages 1065--1069,
  2011.

\bibitem[Glorot and Bengio(2010)]{glorot2010understanding}
X.~Glorot and Y.~Bengio.
\newblock Understanding the difficulty of training deep feedforward neural
  networks.
\newblock In \emph{Aistats}, volume~9, pages 249--256, 2010.

\bibitem[Hinton(2012)]{hinton2012practical}
G.~Hinton.
\newblock A practical guide to training restricted boltzmann machines.
\newblock In \emph{Neural Networks: Tricks of the Trade}, pages 599--619.
  Springer, 2012.

\bibitem[Hinton and Salakhutdinov(2006)]{hinton2006reducing}
G.~Hinton and R.~Salakhutdinov.
\newblock Reducing the dimensionality of data with neural networks.
\newblock \emph{Science}, 313\penalty0 (5786):\penalty0 504--507, 2006.

\bibitem[Kingma and Ba(2014)]{kingma2014adam}
D.~Kingma and J.~Ba.
\newblock Adam: A method for stochastic optimization.
\newblock \emph{arXiv preprint arXiv:1412.6980}, 2014.

\bibitem[Kumar et~al.(2007)Kumar, Chen, and Stern]{kumar2007Profile}
K.~Kumar, T.~Chen, and R.~M. Stern.
\newblock Profile view lip reading.
\newblock In \emph{IEEE International Conference on Acoustics, Speech and
  Signal Processing}, volume~4, pages 429--432, 2007.

\bibitem[Lan et~al.(2012{\natexlab{a}})Lan, Theobald, and
  Harvey]{Lan2012ViewInd}
Y.~Lan, B.~J. Theobald, and R.~Harvey.
\newblock View independent computer lip-reading.
\newblock In \emph{IEEE International Conference on Multimedia and Expo}, pages
  432--437, 2012{\natexlab{a}}.

\bibitem[Lan et~al.(2012{\natexlab{b}})Lan, Theobald, and Harvey]{lam2012}
Y.~Lan, B.~J. Theobald, and R.~Harvey.
\newblock View independent computer lip-reading.
\newblock In \emph{2012 IEEE International Conference on Multimedia and Expo},
  pages 432--437, 2012{\natexlab{b}}.

\bibitem[Lee et~al.(2016)Lee, Lee, and Kim]{lee2016multi}
D.~Lee, J.~Lee, and K.~E. Kim.
\newblock Multi-view automatic lip-reading using neural network.
\newblock In \emph{Workshop on Multi-view Lip-reading Challenges}, pages
  290--302. Asian Conference on Computer Vision, 2016.

\bibitem[Lucey and Potamianos(2006)]{lucey2006ProfVSfrontal}
P.~Lucey and G.~Potamianos.
\newblock Lipreading using profile versus frontal views.
\newblock In \emph{IEEE Workshop on Multimedia Signal Processing}, pages
  24--28, 2006.

\bibitem[Lucey et~al.(2007)Lucey, Potamianos, and Sridharan]{lucey2007Ext}
P.~Lucey, G.~Potamianos, and S.~Sridharan.
\newblock An extended pose-invariant lipreading system.
\newblock In \emph{International Workshop on Auditory-Visual Speech
  Processing}, 2007.

\bibitem[Lucey et~al.(2008)Lucey, Sridharan, and Dean]{lucey2008continuous}
P.~Lucey, S.~Sridharan, and D.~B. Dean.
\newblock Continuous pose-invariant lipreading.
\newblock In \emph{Interspeech}, pages 2679--2682, 2008.

\bibitem[Matthews et~al.(2002)Matthews, Cootes, Bangham, Cox, and
  Harvey]{matthews2002extraction}
I.~Matthews, T.~F. Cootes, A.~Bangham, S.~Cox, and R.~Harvey.
\newblock Extraction of visual features for lipreading.
\newblock \emph{IEEE Transactions on Pattern Analysis and Machine
  Intelligence}, 24\penalty0 (2):\penalty0 198--213, 2002.

\bibitem[Ngiam et~al.(2011)Ngiam, Khosla, Kim, Nam, Lee, and
  Ng]{ngiam2011multimodal}
J.~Ngiam, A.~Khosla, M.~Kim, J.~Nam, H.~Lee, and A.~Y Ng.
\newblock Multimodal deep learning.
\newblock In \emph{Proc. of ICML}, pages 689--696, 2011.

\bibitem[Ninomiya et~al.(2015)Ninomiya, Kitaoka, Tamura, Iribe, and
  Takeda]{ninomiya2015integration}
H.~Ninomiya, N.~Kitaoka, S.~Tamura, Y.~Iribe, and K.~Takeda.
\newblock Integration of deep bottleneck features for audio-visual speech
  recognition.
\newblock In \emph{Conf. of the International Speech Communication
  Association}, 2015.

\bibitem[Petridis and Pantic(2016)]{petridis2016deep}
S.~Petridis and M.~Pantic.
\newblock Deep complementary bottleneck features for visual speech recognition.
\newblock In \emph{IEEE International Conference on Acoustics, Speech and
  Signal Processing}, pages 2304--2308. IEEE, 2016.

\bibitem[Petridis et~al.(2017)Petridis, Li, and
  Pantic]{petridis2017deepVisualSpeech}
S.~Petridis, Z.~Li, and M.~Pantic.
\newblock End-to-end visual speech recognition with lstms.
\newblock In \emph{IEEE International Conference on Acoustics, Speech and
  Signal Processing}, pages 2592--2596. IEEE, 2017.

\bibitem[Potamianos et~al.(2003)Potamianos, Neti, Gravier, Garg, and
  Senior]{Potamianos2003}
G.~Potamianos, C.~Neti, G.~Gravier, A.~Garg, and A.~W. Senior.
\newblock Recent advances in the automatic recognition of audiovisual speech.
\newblock \emph{Proceedings of the IEEE}, 91\penalty0 (9):\penalty0 1306--1326,
  2003.

\bibitem[Saitoh et~al.(2016)Saitoh, Zhou, Zhao, and
  Pietik{\"a}inen]{saitoh2016concatenated}
T.~Saitoh, Z.~Zhou, G.~Zhao, and Matti Pietik{\"a}inen.
\newblock Concatenated frame image based {CNN} for visual speech recognition.
\newblock In \emph{Workshop on Multi-view Lip-reading Challenges}, pages
  277--289. Asian Conference on Computer Vision, 2016.

\bibitem[Sui et~al.(2015)Sui, Togneri, and Bennamoun]{sui2014extracting}
C.~Sui, R.~Togneri, and M.~Bennamoun.
\newblock Extracting deep bottleneck features for visual speech recognition.
\newblock In \emph{IEEE ICASSP}, pages 1518--1522, 2015.

\bibitem[Wand et~al.(2016)Wand, Koutn, and Schmidhuber]{wand2016lipreading}
M.~Wand, J.~Koutn, and J.~Schmidhuber.
\newblock Lipreading with long short-term memory.
\newblock In \emph{IEEE ICASSP}, pages 6115--6119, 2016.

\bibitem[Young et~al.(2002)Young, Evermann, Gales, Hain, Kershaw, Liu, Moore,
  Odell, Ollason, Povey, et~al.]{young2002htk}
S.~Young, G.~Evermann, M.~Gales, T.~Hain, D.~Kershaw, X.~Liu, G.~Moore,
  J.~Odell, D.~Ollason, D.~Povey, et~al.
\newblock The {HTK} book.
\newblock 3:\penalty0 175, 2002.

\bibitem[Zhao et~al.(2009)Zhao, Barnard, and Pietikainen]{zhao2009lipreading}
G.~Zhao, M.~Barnard, and M.~Pietikainen.
\newblock Lipreading with local spatiotemporal descriptors.
\newblock \emph{IEEE Transactions on Multimedia}, 11\penalty0 (7):\penalty0
  1254--1265, 2009.

\bibitem[Zimmermann et~al.(2016)Zimmermann, Ghazi, Ekenel, and
  Thiran]{zimmermann2016visual}
M.~Zimmermann, M.~M. Ghazi, H.~K. Ekenel, and J.~P. Thiran.
\newblock Visual speech recognition using {PCA} networks and {LSTMs} in a
  tandem {GMM-HMM} system.
\newblock In \emph{Asian Conference on Computer Vision}, pages 264--276.
  Springer, 2016.

\end{thebibliography}
\end{document}